\documentclass[10pt,twocolumn,letterpaper]{article}

\usepackage[pagenumbers]{cvpr} 

\definecolor{cvprblue}{rgb}{0.21,0.49,0.74}
\usepackage[pagebackref,breaklinks,colorlinks,allcolors=cvprblue]{hyperref}
\usepackage{hyperref, url, algorithm, algpseudocode, multirow, adjustbox, tabularx, booktabs, amsmath, amssymb, pifont, mathtools}
\usepackage{CJKutf8}    
\usepackage{colortbl}
\usepackage[table]{xcolor}
\usepackage[misc]{ifsym}        

\newcolumntype{G}{>{\columncolor{gray!12}}S[table-format=3.2]}
\newcolumntype{N}{S[table-format=3.2]}

\definecolor{headerblue}{RGB}{70, 130, 180}
\definecolor{lightblue}{RGB}{230, 240, 250}
\definecolor{bestcolor}{RGB}{255, 248, 220}
\definecolor{humancolor}{RGB}{245, 245, 245}
\definecolor{accentcolor}{RGB}{255, 235, 205}

\def\confName{CVPR}
\def\confYear{2026}

\newcommand{\model}{WAM-Flow}

\title{\model: Parallel Coarse-to-Fine Motion Planning via Discrete Flow Matching for Autonomous Driving}

\newcommand{\authorskipshort}{\hspace{3mm}}
\author{
  \begin{tabular}{c}
  Yifang~Xu$^{1*}$ \authorskipshort
  Jiahao~Cui$^{1*}$ \authorskipshort
  Feipeng~Cai$^{2*}$ \authorskipshort
  Zhihao~Zhu$^{1}$ \authorskipshort
  Hanlin~Shang$^{1}$ \authorskipshort
  Shan~Luan$^{1}$
  \\ [1mm]
  Mingwang~Xu$^{1}$ \authorskipshort
  Neng~Zhang$^2$ \authorskipshort
  Yaoyi~Li$^2$  \authorskipshort
  Jia~Cai$^2$  \authorskipshort
  Siyu~Zhu$^{1}\textsuperscript{\Letter}$
  \end{tabular}
  \\ [4mm]
  $^1$Fudan University \hspace{8mm}
  $^2$Yinwang Intelligent Technology Co., Ltd
  \\ [1mm]
  \normalsize{Code \& Model: \url{https://github.com/fudan-generative-vision/WAM-Flow}}
}

\begin{document}
\maketitle

\begingroup
\renewcommand\thefootnote{}\footnote{
    $^*$: Equal contribution.  \hspace{18mm}   \Letter: Corresponding authors.
} \\
\renewcommand\thefootnote{}\footnote{
    \{xuyf25, cuijh25\}@m.fudan.edu.cn  
    \quad  siyuzhu@fudan.edu.cn
}
\endgroup

\begin{abstract}
We introduce \model, 
a vision–language–action (VLA) model that casts ego-trajectory planning as discrete flow matching over a structured token space. 
In contrast to autoregressive decoders, \model\space performs fully parallel, bidirectional denoising, enabling coarse-to-fine refinement with a tunable compute–accuracy trade-off. 
Specifically, the approach combines a metric-aligned numerical tokenizer that preserves scalar geometry via triplet-margin learning, 
a geometry-aware flow objective and a simulator-guided GRPO alignment that integrates safety, ego progress, and comfort rewards while retaining parallel generation. 
A multi-stage adaptation converts a pre-trained auto-regressive backbone (Janus-1.5B) from causal decoding to non-causal flow model and strengthens road-scene competence through continued multimodal pretraining.
Thanks to the inherent nature of consistency model training and parallel decoding inference, 
\model\space achieves superior closed-loop performance against autoregressive and diffusion-based VLA baselines, 
with 1-step inference attaining 89.1 PDMS and 5-step inference reaching 90.3 PDMS on NAVSIM v1 benchmark. 
These results establish discrete flow matching as a new promising paradigm for end-to-end autonomous driving. 
\textbf{The code will be publicly available soon.}
\end{abstract}

\begin{figure*}[t!]
  \centering
  \includegraphics[width=\textwidth]{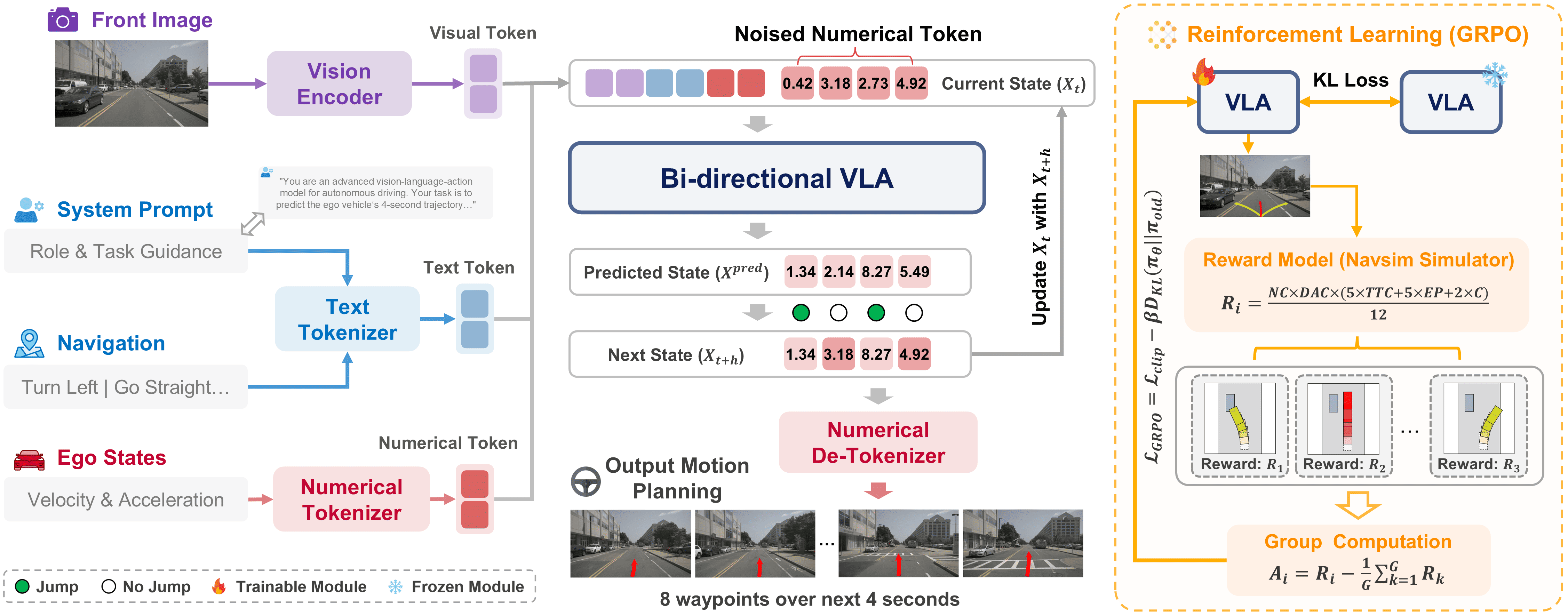} 
  \vspace{-6mm}
  \caption{Architecture of the proposed \model\space framework. Our method takes as input a front-view image, a natural-language navigation command with a system prompt, and the ego-vehicle states, and outputs an 8-waypoint future trajectory spanning 4 seconds through parallel denoising. The model is first trained via supervised fine-tuning to learn accurate trajectory prediction. We then apply simulator-guided GRPO to further optimize closed-loop behavior. The GRPO reward function integrates safety constraints (collision avoidance, drivable-area compliance) with performance objectives (ego-progress, time-to-collision, comfort).}
  \label{fig:pipeline}
\end{figure*}

\section{Introduction}
Vision-language–action models for end-to-end autonomous driving~\cite{zhou2025autovla, FutureSightDrive-2025, RecogDrive-2025} aim to map egocentric driving-view video inputs and natural-language instructions into both causal reasoning and precise ego-vehicle motion planning, while satisfying stringent efficiency and safety requirements.
A fundamental challenge in this domain is the design of a policy representation that effectively balances three critical aspects: expressive reasoning capabilities, high-fidelity continuous control, and robust closed-loop performance.
Existing approaches can be broadly categorized into dual-system and single-system paradigms.
Dual-system methods~\cite{RecogDrive-2025, Epona-2025, AdaDrive-2025, Alpamayo-R1-2025, DriveAgent-2025} typically employ autoregressive vision-language models (VLMs)~\cite{LLaMA-2, LLaVA, wang2024qwen2, Qwen-25, InternVL3-2025} as auxiliary reasoning modules to provide high-level driving intent, scene summaries, or linguistic guidance for downstream motion planning networks, 
which often utilize diffusion-based iterative optimization~\cite{DDPM-2020, DiffusionDrive-2025, jiang2025diffvla, Artemis-2025} to generate smooth, complex action distributions.
In contrast, single-system approaches~\cite{EMMA-2024, OpenEMMA-2025, DrivingGPT-2025, FutureSightDrive-2025, zhou2025autovla} such as EMMA~\cite{EMMA-2024} and DrivingGPT~\cite{DrivingGPT-2025} reformulate trajectory or action prediction as a text generation problem within the VLM, enabling reasoning and planning directly in the linguistic space.
This work investigates a novel alternative based on discrete flow matching (DFM), 
which offers distinct advantages for autonomous driving applications.

Discrete flow matching~\cite{DFM-2024, DFM-2025, wang2025fudoki, deng2025uniform, su2025theoretical, karimi2025fs, yue2025oat, cheng2025alpha} models probability transport over discrete token spaces via a continuous-time Markov chain (CTMC) that carries a simple base distribution to the data distribution.
Unlike autoregressive decoders that commit to tokens sequentially and accumulate exposure-bias errors, 
Discrete flow matching supports fully parallel denoising and bidirectional refinement during generation. 
These properties enable coarse-to-fine planning: 
beginning with a coarse motion hypothesis, 
the model increases trajectory fidelity through additional denoising steps, 
yielding a tunable compute–accuracy trade-off. 
This flexibility aligns well with autonomous driving, 
where simple scenes admit rapid approximate plans while complex interactions require higher-precision refinement. 
Despite these advantages, discrete flow matching remains largely unexplored for VLA policies in end-to-end autonomous driving.

However, a straightforward application of discrete flow matching to VLA model for end-to-end autonomous driving is nontrivial for three reasons. 
First, training discrete flow matching from scratch is prohibitively data- and compute-intensive, 
so they are typically initialized from general-purpose autoregressive multimodal VLMs that lack sufficient road-scene competence--from low-level perception and motion forecasting to high-level planning and decision making. 
We therefore adopt a multi-stage adaptation strategy: 
starting from a generic VLM backbone (Janus-1.5B~\cite{Janus-2025}), 
we continued conduct pretraining on large-scale road-scene visual question answering (VQA) to strengthen the ability to understand various complex road scenes and vehicle driving patterns, establishing a strong domain prior comparable to autoregressive VLA baselines.
Second, standard text token embeddings are ill-suited to high-precision numerical regression because they weakly encode metric relationships. 
We introduce a metric-aligned numerical tokenizer that discretizes continuous scalars into a shared codebook and learns embeddings with a triplet-margin ranking objective so that latent distances reflect underlying scalar differences. 
This structured token space enables stable coarse-to-fine and slow–fast trajectory refinement within discrete flow matching, 
providing a controllable compute–accuracy trade-off.
Finally, supervised likelihood-based flow training aligns the model with expert trajectories but does not explicitly enforce safety, ego-progress, and comfort in closed-loop control. 
We incorporate a Group Relative Policy Optimization (GRPO) based alignment objective with a composite reward that integrates safety penalties and performance goals, 
improving the safety–progress–comfort profile while preserving the model’s parallel generation capabilities.

Experimental results on the NAVSIM v1 and v2 benchmarks demonstrate that \model\space achieves superior performance in PDMS and EPDMS metrics compared to both autoregressive and diffusion-based VLA models. 
By leveraging discrete flows over a structured token space, 
\model\space enables flexible slow–fast and coarse-to-fine trajectory prediction. 
With 1-step denoising, 
it attains competitive performance (89.1 PDMS), 
while 5-step refinement yields further gains (90.3 PDMS). 
On the NAVSIM v2 benchmark, the full model achieves 84.7 EPDMS. 
Notably, the 1.5B-parameter \model\space model achieves an 3$\times$ improvement in inference speed over the Janus autoregressive baseline, 
underscoring the promising effectiveness and efficiency of the discrete flow matching approach for end-to-end autonomous driving.

\section{Related Work}
\noindent\textbf{VLMs in Autonomous Driving.}
Autoregressive VLMs ~\cite{EMMA-2024, OpenEMMA-2025, FutureSightDrive-2025, zhou2025autovla, tian2024drivevlm, jiang2024senna} formulate driving as a sequential language modeling problem, 
where each token corresponds to a trajectory point, control command, or reasoning step.
Representative works such as EMMA~\cite{EMMA-2024}, OpenEMMA~\cite{OpenEMMA-2025}, FutureSightDrive~\cite{FutureSightDrive-2025} and AutoVLA~\cite{zhou2025autovla}
leverage chain-of-thought reasoning and external memory modules to enhance interpretability and decision transparency.
Despite their strong causal modeling capability, 
autoregressive architectures suffer from slow autoregressive decoding and limited parallelism, 
as future actions must be generated step-by-step.
Diffusion-based methods, including DiffusionDrive~\cite{DiffusionDrive-2025}, ViLaD~\cite{cui2025vilad} and DiffVLA~\cite{jiang2025diffvla}
treat planning as a denoising process that gradually refines latent trajectory representations. 
These models enable parallel sampling but often lack explicit reasoning interpretability.
In this paper, we explore a new promising paradigm for end-to-end autonomous driving, namely discrete flow matching.

\noindent\textbf{Discrete Diffusion in LLMs and VLMs.}
Recent progress in discrete generative modeling has led to the emergence of discrete diffusion LLMs~\cite{nie2025llada,ye2025dream} 
and discrete diffusion VLMs~\cite{nie2025large,yu2025dimple,you2025llada-v,li2025lavida,yang2025mmada}, 
which extend diffusion processes to tokenized sequences. 
This direction originates from D3PM~\cite{austin2021-d3pm}, 
which formulated diffusion as a discrete Markov process over categorical variables.
Recently, LLaDA~\cite{nie2025llada} trains an 8B-parameter model from scratch, 
reaching LLaMA-3~\cite{grattafiori2024llama-3} performance with bidirectional reasoning and robustness. 
DREAM-7B~\cite{ye2025dream} further enhances diffusion-based reasoning via iterative refinement and arbitrary-order generation.
Meanwhile, discrete flow matching~\cite{gat2024discrete,lipman2024flow,shaul2024flow} generalizes discrete diffusion via continuous-time probability paths and learnable velocity fields. 
By unifying diffusion and flow-based generation under a single probabilistic framework, 
it enables parallel, bidirectional, and efficient sampling. 
FUDOKI~\cite{wang2025fudoki} extends this framework to multimodal reasoning and generation,
demonstrating unified, non-autoregressive modeling across modalities.
In this paper, we apply discrete flow matching to VLA for autonomous driving and explore its inherent nature of parallel generation and coarse-to-fine controllability.

\noindent\textbf{Reinforcement Learning in VLA.}
Building upon the success of DeepSeek-R1~\cite{guo2025deepseek}, 
GRPO has been further extended to autonomous driving domains.
In particular, AlphaDrive~\cite{jiang2025alphadrive} pioneers the integration of GRPO-based reinforcement learning with planning-centric reasoning in autonomous driving, achieving notable improvements in both decision-making performance and training efficiency.
TrajHF~\cite{li2025finetuning} further combines diffusion-based multimodal planners with reinforcement learning from human feedback, 
enabling safe and personalized trajectory generation aligned with diverse human driving styles. 
More recently, AutoVLA~\cite{zhou2025autovla} incorporates GRPO into vision-language-action models, 
extending reinforcement learning to end-to-end multimodal reasoning and low-level planning. 
To the best of our knowledge, this work presents the first exploration of GRPO within discrete flow matching for autonomous driving VLA.
Furthermore, we explicitly incorporate safety alignment objectives, extending beyond conventional likelihood-based training to enhance reliability in autonomous driving contexts.
\vspace{-0.2cm}
\section{Method}
\label{sec:method}
We present \model, a VLA model that formulates motion planning as a discrete flow matching problem over a structured token space. 
Specifically, Section~\ref{subsec:preliminary} establishes the theoretical foundation of discrete flow matching over finite alphabets. 
Building on this, Section~\ref{subsec:architecture} details the model architecture, 
including a metric-aligned numerical tokenizer, and a geometry-aware flow objective. 
To address the limitations of likelihood-based training, Section~\ref{subsec:grpo} introduces simulator-guided GRPO to enforce safety and performance in closed-loop control. 
Finally, Section~\ref{subsec:train_infer} specifies the autoregressive-to-flow training and the parallel denoising–based inference.
Figure~\ref{fig:pipeline} demonstrates the pipeline of \model.

\subsection{Preliminaries: Discrete Flow Matching}
\label{subsec:preliminary}
\noindent\textbf{Probability Paths.}
Let the discrete state space be defined as $S = \mathcal{T}^D$, 
where $\mathcal{T} = [K] = \{1, \dots, K\}$ represents a set of possible discrete values, 
and $D$ is the number of discrete variables. 
Denote the data distribution by $q(x)$ over $S$ and a simple factorized source distribution by $p(x) = \prod_{i=1}^D p^i(x^i)$. 
We define a time-dependent probability path $\{p_t(x)\}_{t \in [0,1]}$ by marginalizing conditional, 
coordinate-wise factorized paths around a latent target $x_1$:
\begin{equation}
\small
p_t(x) = \sum_{x_1 \in S} q(x_1)\, p_t(x | x_1), 
\; 
p_t(x | x_1) = \prod_{i=1}^D p_t^i\big(x^i | x_1^i\big),
\end{equation}
with boundary conditions ensuring 
$p_0^i(\cdot | x_1^i) = p^i(\cdot)$ 
and 
$p_1^i(\cdot | x_1^i) = \delta_{x_1^i}(\cdot)$, 
which yields $p_0(x) = p(x)$ and $p_1(x) = q(x)$. 
This mixture construction separates the definition of the transport path from the generative dynamics. 
A common instance is the mixture (mask) path:
\begin{equation}
p_t^i(x^i | x_1^i) = \big(1 - \kappa_t\big)\, p^i(x^i) + \kappa_t\, \delta_{x_1^i}(x^i),
\end{equation}
where $\kappa_t \in [0,1]$ is a monotonically increasing scheduling function satisfying $\kappa_0 = 0$ and $\kappa_1 = 1$. 
When $p^i(x^i) = \delta_{\text{[MASK]}}(x^i)$, this path recovers the standard masked corruption process.

\noindent\textbf{Generative Dynamics.}
The probability path $p_t(x)$ is realized through a CTMC characterized by a probability velocity $u_t(x, z)$. 
This velocity acts as a rate matrix, defining the instantaneous transition rate from state $z$ to state $x$ at time $t$. 
Formally, for a small time step $h > 0$, the transition probability satisfies:
\begin{equation}
P(x_{t+h} = x | x_t = z) = \delta_z(x) + h u_t(x, z) + o(h),
\end{equation}
where $\delta_z(x)$ is the Kronecker delta and $o(h)$ denotes higher-order terms. 
The velocity $u_t$ must adhere to the constraints: 
$u_t(x, z) \geq 0$ for all $x \neq z$, and $\sum_x u_t(x, z) = 0$. 
This velocity generates the path $p_t$ via the Kolmogorov forward equation:
\begin{equation}
\dot{p}_t(x) + \operatorname{div}_x(j_t) = 0,
\end{equation}
where the probability flux is given by $j_t(x, z) = u_t(x, z) p_t(z)$. 
To maintain tractability in high-dimensional spaces, we restrict the velocity to permit only single-coordinate transitions.

\subsection{\model\space Architecture}
\label{subsec:architecture}
\noindent\textbf{Problem Formulation.}
We formulate the motion planning task as a conditional sequence generation problem. 
The model maps multimodal inputs--including synchronized front-view camera images, 
a natural-language navigation command, 
and the current ego-vehicle state (position, heading, velocity and acceleration)--to a discrete token sequence representing the planned trajectory. 
The output is a sequence of 8 waypoints spanning the next 4 seconds.

Within this formulation, 
\model\space employs a flow network that learns to transport a simple prior distribution over the discrete token space to the expert trajectory distribution. 
An advantage of this approach is its support for fully parallel token transitions during generation, 
which circumvents the sequential bottleneck of autoregressive decoding. 
This capability enables a flexible trade-off between computational efficiency and prediction fidelity: 
rapid, coarse plans can be generated with few denoising steps, while high-precision trajectories are achieved through iterative refinement.

\noindent\textbf{Metric-Aligned Numerical Tokenizer.}
Standard text token embeddings do not preserve metric structure and thus perform poorly for high-precision regression. 
We introduce a metric-aligned numerical tokenizer that discretizes continuous scalars (e.g., position, heading, velocity and acceleration) into a uniform codebook 
$\mathcal{V}=\{v_1,\dots,v_N\}$ 
over $[-100,100]$ with 0.01 resolution ($N=20{,}001$). 
Each scalar token $v$ is mapped by a linear projection 
$E:\mathbb{R}\to\mathbb{R}^d$ and L2-normalized to yield the embedding $z=E(v)/\lVert E(v)\rVert_2$.

To align latent geometry with numeric distances, we enforce that Euclidean embedding distances are monotonic in the underlying scalar differences. 
Let $d_{ij}=\lVert z_i-z_j\rVert_2$. 
For any triplet $(i,j,k)$ with $|v_i-v_j|<|v_i-v_k|$, 
we promote $d_{ij}<d_{ik}$ via a triplet-margin ranking loss:
\begin{equation}
\label{eq:num_loss}
\mathcal{L}_{\mathrm{num}}
=\mathbb{E}_{(i,j,k)\sim\mathcal{T}}\big[\max\big(0,\; d_{ij}-d_{ik}+\alpha\big)\big],
\end{equation}
where $\mathcal{T}$ samples anchors $i$ with near/far neighbors $(j,k)$ and $\alpha>0$ is a fixed margin. 
This construction yields a numerically coherent token space in which latent distances faithfully reflect scalar proximity, 
enabling stable coarse-to-fine and slow–fast refinement under discrete flow matching. 
The induced distances serve as the tokenizer-specific metric $d_i(\cdot,\cdot)$ in the geometry-aware flow objective.

\noindent\textbf{Discrete Flow Matching Objective.}  
To respect the geometric structure of the tokenized action space, 
we design a conditional probability path that is both tractable and expressive. 
Given a target sequence $x_1 \in q(x)$, 
we define a Gibbs distribution induced by a distance metric $d$:  
\begin{equation}
\small
p_t(x | x_1) = \mathrm{softmax}\left(-\beta_t d(x, x_1)\right), \; 
\beta_0 = 0, \; 
\beta_1 \to \infty,
\end{equation}  
where $\beta_t$ is a monotonically increasing scheduling function on $[0,1]$, 
and $d(x, x_1) = \sum_{i=1}^{D} w_i d_i(x^i, x_1^i)$ is a weighted sum of coordinate-wise dissimilarities. 
Each $d_i$ is tailored to the data type: 
tokenizer-induced distances for numerical values, 
circular metrics for angles, 
and semantic distances for textual fields. 
The nonnegative weights $w_i$ balance the contribution of each coordinate.

This path is realized by a CTMC with a transition rate designed to steer the state toward the target. 
The conditional rate for transitioning from $z$ to $x$ given $x_1$ is:  
\begin{equation}
\label{eq:flow_trans}
u_t(x, z | x_1) = p_t(x | x_1) \dot{\beta}_t \left[d(z, x_1) - d(x, x_1)\right]_+,
\end{equation}
where $[ \cdot ]_+ = \max(0, \cdot)$. 
This rate assigns higher probability to transitions that reduce the dissimilarity to the target. 
The marginal velocity is obtained by integrating over the posterior distribution of $x_1$ given the current state.

The model is trained to approximate the true posterior $p_{1|t}(x_1 | x)$ by minimizing the conditional flow matching cross-entropy loss:  
\begin{equation}
\small
\label{eq:flow_loss}
\mathcal{L}_{\mathrm{CE}}(\theta) = \mathbb{E}_{t \sim \mathcal{U}[0,1],\, x_1 \sim q,\, x \sim p_t(\cdot | x_1)} \left[ -\sum_{i=1}^{D} \log p^{\theta,i}_{1 | t}(x_1^i
 |x) \right],
\end{equation}  
where $p_{1 | t}^{\theta,i}(x_1^i | x)$ is the model’s estimate of the posterior probability for the $i$-th target token. 
This geometry-aware formulation enables efficient parallel decoding and supports controllable refinement, 
allowing flexible trade-offs between planning speed and trajectory quality.

\noindent\textbf{Model Architecture.}
We adapt a Janus-1.5B multimodal backbone to the discrete flow matching generation paradigm for vision–language–action planning. 
Images are resized with preserved aspect ratio, zero-padded to 384×384, 
and encoded by SigLIP~\cite{SigLIP-2023} into 576 visual tokens; 
a lightweight MLP aligns these features to the 2048-dimensional Janus text-token space. 
On the language side, 
we extend the Janus tokenizer by 20,001 numerically grounded tokens to represent input ego-state numbers and output waypoint coordinates, 
yielding a 122,401-word vocabulary. 
Training data are formatted with a fixed QA-style prompt that integrates navigation commands, ego-state (position, heading, velocity, acceleration), 
and the target waypoint sequence for the next 4 seconds (8 waypoints). 
For the decoder,
the original Janus text head is expanded to the enlarged vocabulary and used to predict action tokens under the discrete flow matching objective.



\subsection{Simulator-Guided GRPO}
\label{subsec:grpo}
While supervised flow matching optimizes trajectory prediction accuracy, 
it does not explicitly enforce critical driving objectives such as safety, comfort, and progress in closed-loop control. 
To address this limitation, 
we introduce an online GRPO reinforcement learning that aligns the policy with simulator-derived rewards while preserving the parallel generation capabilities of discrete flow matching.

\noindent\textbf{Reward Design.}
We design a composite reward function that decomposes the NAVSIM simulator's PDMS metrics into safety penalties and performance objectives. 
The reward for a generated trajectory $\tau$ is defined as:
\begin{equation}
R(\tau) = \underbrace{\left( \prod_{m \in \mathcal{M}} s_m(\tau) \right)}_{\text{safety penalties}} \cdot \underbrace{\left( \frac{ \sum_{w \in \mathcal{W}} \lambda_w s_w(\tau) }{ \sum_{w \in \mathcal{W}} \lambda_w } \right)}_{\text{performance objectives}},
\end{equation}
where $\mathcal{M} = \{\mathrm{NC}, \mathrm{DAC}\}$ 
represents safety metrics, including no-collision and drivable-area compliance;
$\mathcal{W} = \{\mathrm{EP}, \mathrm{TTC}, \mathrm{C}\}$ 
denotes performance metrics, including ego-progress, time-to-collision, and comfort. 
The multiplicative safety term ensures strict constraint satisfaction, 
while the weighted average balances performance trade-offs. 
Specifically, the NC score assigns $s_{\mathrm{NC}}(\tau) = 0$ for at-fault collisions, 0.5 for collisions with static objects, and 1 otherwise; 
DAC yields $s_{\mathrm{DAC}}(\tau) = 0$ on violations and 1 otherwise. 
Sub-scores $s_w(\tau) \in [0,1]$ are normalized, 
with $\lambda_w \geq 0$ as weighting coefficients.

\vspace{1mm}
\noindent\textbf{GRPO Objective.}
For a given scene context c, 
we sample $G$ candidate trajectories $\{\tau_i\}_{i=1}^G \sim \pi_{\theta_{\mathrm{old}}}(\cdot | c)$ via parallel denoising. 
Each trajectory receives a reward $R_i = R(\tau_i)$. 
Using the group baseline $A_i = R_i - \frac{1}{G}\sum_{j=1}^G R_j$, 
we define per-token importance ratios for action tokens $\{o_i^k\}_{k=1}^{T_i}$ under conditioning states $\{s_i^k\}$:
$r_i^k(\theta) = \frac{\pi_{\theta}(o_i^k s_i^k)}{\pi_{\theta_{\mathrm{old}}}(o_i^k
 s_i^k)}$.
The GRPO surrogate objective, with clipping parameter $\epsilon > 0$ and KL regularization strength $\beta \geq 0$, is formulated as:
%
%
{\scriptsize
\begin{equation}
\label{eq:grpo_loss}
\begin{aligned}
\mathcal{L}_{\mathrm{GRPO}}(\theta) = \mathbb{E}_{c} \Bigg[ \frac{1}{G} \sum_{i=1}^G \frac{1}{T_i} \sum_{k=1}^{T_i} \Big( & \min\left\{ r_i^k(\theta) A_i, \mathrm{clip}(r_i^k(\theta), 1-\epsilon, 1+\epsilon) A_i \right\} \\
& - \beta D_{\mathrm{KL}}\left( \pi_\theta(\cdot s_i^k) \ \pi_{\mathrm{ref}}(\cdot s_i^k) \right) \Big) \Bigg].
\end{aligned}
\end{equation}
}
The group baseline reduces variance by inducing relative preferences within each sample set, 
while the KL divergence term stabilizes updates by anchoring the policy to the supervised reference.

\begin{figure}[t!]
  \centering
  \includegraphics[width=\linewidth]{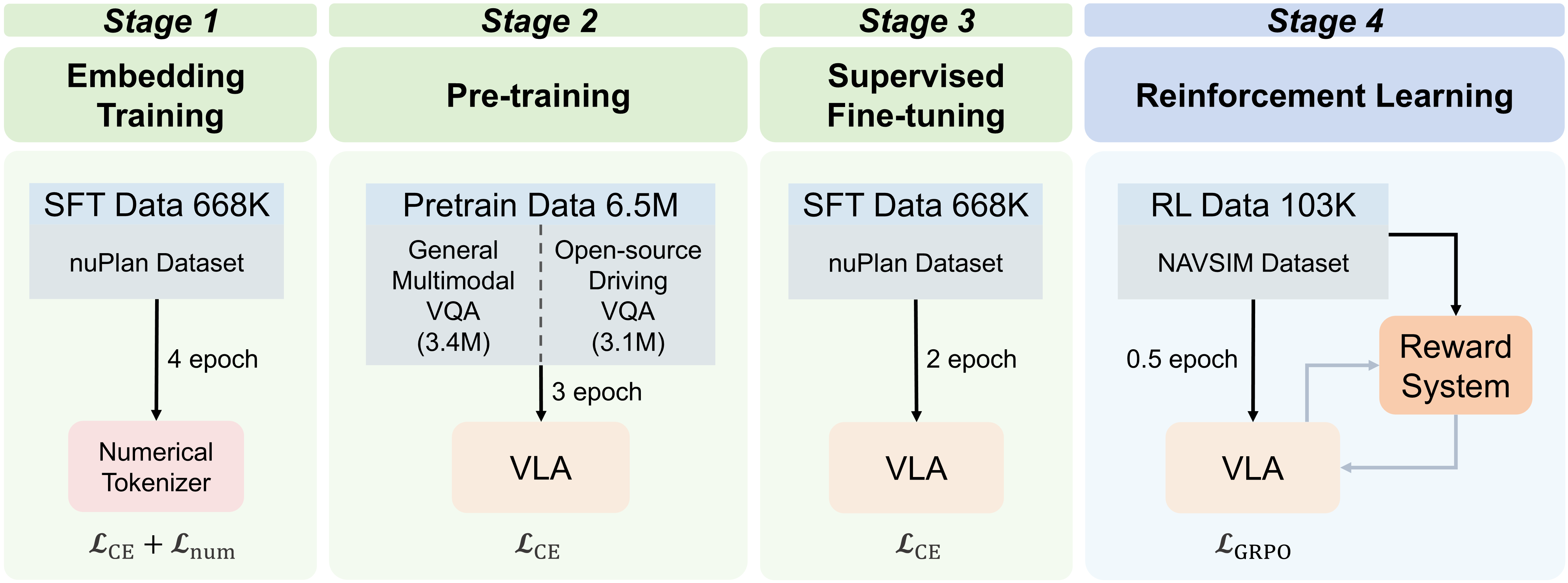} 
  \vspace{-6mm}
  \caption{Overview of the full training curriculum.
  Different training stage motivation and corresponding training data and training steps are demonstrated.}
  \label{fig:training}
\end{figure}

\subsection{Training and Inference}
\label{subsec:train_infer}

\noindent\textbf{Autoregressive-to-Flow Training.} 
Figure~\ref{fig:training} outlines a four-stage curriculum. 
First, we randomly initialize the numerical embeddings and freeze the VLA backbone. We train the numerical embeddings together with the language-model head on the 668K nuPlan dataset for 4 epochs, using flow-matching loss $\mathcal{L}_{\mathrm{CE}}$ (Equation~\ref{eq:flow_loss}) and triplet-margin ranking loss $\mathcal{L}_{\mathrm{num}}$ (Equation~\ref{eq:num_loss}). 
Second, we enhance the perception of driving scenes by pretraining VLA using $\mathcal{L}_{\mathrm{CE}}$.
This stage trains 3 epochs on 6.5M VQA, 
including general multimodal VQA (3.4M) from LLaVA-v1.5~\cite{LLaVA} and large-scale driving-specific VQA (3.1M) from RecogDrive~\cite{RecogDrive-2025}, 
which enhances perceptual grounding and driving-specific causal reasoning. 
Third, we supervised fine-tuning of the VLA backbone for only 2 epochs on the nuPlan dataset with $\mathcal{L}_{\mathrm{CE}}$. 
After supervised flow training, 
we perform reinforcement learning with simulator feedback by maximizing the GRPO objective (Equation~\ref{eq:grpo_loss}) with KL regularization toward the supervised reference to optimize our VLA model for 0.5 epoch on 103k NAVSIM dataset. 
We set the weight for $\mathrm{EP}, \mathrm{TTC}$ and $\mathrm{C}$ in our reward to 5:5:2.



\noindent\textbf{Inference.}
First, we apply the Euler discretization over the time interval $[0, 1]$ with $n$ inference steps, yielding a step size of $h = \frac{1}{n}$. 
For each coordinate $i$, the initial token $x_0^i$ is sampled uniformly from the model vocabulary. 
At each discrete timestep $t \in [0, 1]$, the current token is denoted as $x_t^i$, and a target token $x_1^i$ is drawn from the posterior distribution $p_{1 | t}^{i}(x_1^i | x)$.
Next, we compute the total outgoing transition rate $\lambda_i$ for the current token $x_t^i$ as
$\lambda_i = \sum_{x^i \neq x_t^i} u_t^i(x^i, x_t^i| x_1^i)$, where the conditional rate function $u_t^i$ is defined in Equation~\ref{eq:flow_trans}. 
A uniform random variable $Z_i \sim  \mathcal{U}[0,1]$ is then drawn.
The jump rule is as follows: if $Z_i < 1 - e^{-h \lambda_i}$, a transition occurs, and the new token $x_{t+h}^i$ is sampled proportionally to the normalized rates $u_t^i(\cdot, x_t^i| x_1^i)$; otherwise, the token remains unchanged, i.e., $x_{t+h}^i = x_t^i$. 
After $n$ sampling steps, we obtain the final output token sequence $x_1$.

\begin{table*}[!t]
\centering
\begin{minipage}[t]{0.57\textwidth}
    \centering
    \small
    \setlength{\tabcolsep}{4pt}
    \resizebox{1.11\linewidth}{!}{
    \begin{tabular}{lccccccccc}
    \toprule
    \textbf{Method} & \textbf{Paradigm} & \textbf{Backbone} & \textbf{Input} & \textbf{NC}$\uparrow$ & \textbf{DAC}$\uparrow$ & \textbf{TTC}$\uparrow$ & \textbf{Comf.}$\uparrow$ & \textbf{EP}$\uparrow$ & \textbf{PDMS}$\uparrow$\\
    \midrule
    \textit{End-to-End} \\
    VADv2~\cite{VADv2-2024} & - & - & 6$\times$Cam & 97.2 & 89.1 & 91.6 & \textbf{100} & 76.0 & 80.9 \\
    Transfuser~\cite{Transfuser-2022} & - & - & 3$\times$Cam + L& 97.7 & 92.8 & 92.8 & \textbf{100} & 79.2 & 84.0 \\
    Hydra-MDP++~\cite{Hydra-MLP-pp-2025} & - & - & 3$\times$Cam + L & 98.3 & 96.0 & 94.6 & \textbf{100} & 78.7 & 86.5\\
    Artemis~\cite{Artemis-2025} & Diff. & - & 6$\times$Cam & 98.3 & 95.1 & 94.3 & 99.8 & 81.4 & 87.0\\
    DiffusionDrive~\cite{DiffusionDrive-2025} & Diff. & - & 3$\times$Cam + L & 98.2 & 96.0 & 94.8 & \textbf{100} & 82.2 & 88.1 \\
    \midrule
    \textit{End-to-End VLA} \\
    DrivingGPT~\cite{DrivingGPT-2025} & AR & LLaMA2-7B~\cite{LLaMA-2} & 1$\times$Cam & 98.1 & 90.7 & 94.9 & 95.6 & 79.7 & 82.4 \\
    FSDrive~\cite{FutureSightDrive-2025} & AR & Qwen2-VL-2B~\cite{wang2024qwen2} & 6$\times$Cam & 98.2 & 93.8 & 93.3 & 99.9 & 80.1 & 85.1\\
    Epona~\cite{Epona-2025} & AR + Diff. & DiT-2.5B~\cite{DiT-2023} & 1$\times$Cam & 97.9 & 95.1 & 93.8 & 99.9 & 80.4 & 86.2 \\
    AutoVLA~\cite{zhou2025autovla} & AR & Qwen2.5-3B~\cite{Qwen-25} & 3$\times$Cam & 98.4 & 95.6 & \textbf{98.0} & 99.9 & 81.9 & 89.1\\
    ReCogDrive~\cite{RecogDrive-2025} & AR + Diff. & InternVL3-8B~\cite{InternVL3-2025} & 3$\times$Cam & 98.2 & 97.5 & 95.2 & 99.9 & \textbf{83.5} & 89.6\\
    \textbf{Ours} & DFM & Janus-1.5B~\cite{Janus-2025} & 1$\times$Cam & \textbf{99.2} & \textbf{98.3} & 97.0 & 99.7 & 82.3 & \textbf{90.3} \\
    \midrule
    \end{tabular}}
    \vspace{-3mm}
    \caption{Comparison on NAVSIM-v1 with closed-loop metrics. Abbreviation: Diff.(Diffusion), Comf.(Comfort), Cam (Camera), L (LiDAR).
    }
    \label{tab:navsim_v1}
\end{minipage}
\hfill
\begin{minipage}[t]{0.37\textwidth}
    \centering
    \small
    \vspace{-2.16cm}
    \setlength{\tabcolsep}{4pt}
    \resizebox{0.95\linewidth}{!}{
    \begin{tabular}{c|cccccc}
    \toprule
    \textbf{Group Size} & NC$\uparrow$ & DAC$\uparrow$ & TTC$\uparrow$ & Comf.$\uparrow$ & EP$\uparrow$ & \textbf{PDMS}$\uparrow$\\
    \midrule
    w/o GRPO & 98.5  & 95.1  & 94.4  & 99.5 & 81.8 & 86.7 \\
    2&  \textbf{99.4} & 97.3 &  96.8 &  99.7 & 80.7 & 89.2 \\
    3& 99.2 & \textbf{98.3} & \textbf{97.0} & 99.7 & \textbf{82.3} & \textbf{90.3} \\
    4&  99.3 & 97.6 &  96.5 &  \textbf{99.8} & 82.0 & 89.6 \\
    \bottomrule
    \end{tabular}
    }
    \vspace{-3mm}
    \caption{Ablation on GRPO group size.}
    \vspace{2mm}
    \label{tab:ablation_group_size}

    \centering
    \small
    \setlength{\tabcolsep}{4pt}
    \resizebox{0.95\linewidth}{!}{
    \begin{tabular}{c|cccccc}
    \toprule
    EP : TTC : C & NC$\uparrow$ & DAC$\uparrow$ & TTC$\uparrow$ & Comf.$\uparrow$ & EP$\uparrow$ & \textbf{PDMS}$\uparrow$\\
    \midrule
    5:20:2 &\textbf{99.5} &\textbf{98.3} &\textbf{97.9}  &99.6 &80.1 &89.7 \\
    5:5:8 &99.4 &98.1 &96.9  &\textbf{99.7}  &82.1 &90.1 \\
    20:5:2&99.4 &98.1  &96.4 &99.3   &\textbf{82.7} &90.0 \\
    \midrule
    5:5:2 & 99.2 & \textbf{98.3} & \textbf{97.0} & \textbf{99.7} & 82.3 & \textbf{90.3}  \\
    \bottomrule
    \end{tabular}
    }
    \vspace{-3mm}
    \caption{Ablation on different weight of Simulator-Guided reward.
    The default weight is 5:5:2 for Navsim simulator, 
    and we adjust the scale of each weight by $4\times$ to obtain the new weight.}
    \label{tab:ablation_reward}
\end{minipage}
\end{table*}

\begin{figure*}[t!]
  \vspace{1mm}
  \centering
  \includegraphics[width=\linewidth]{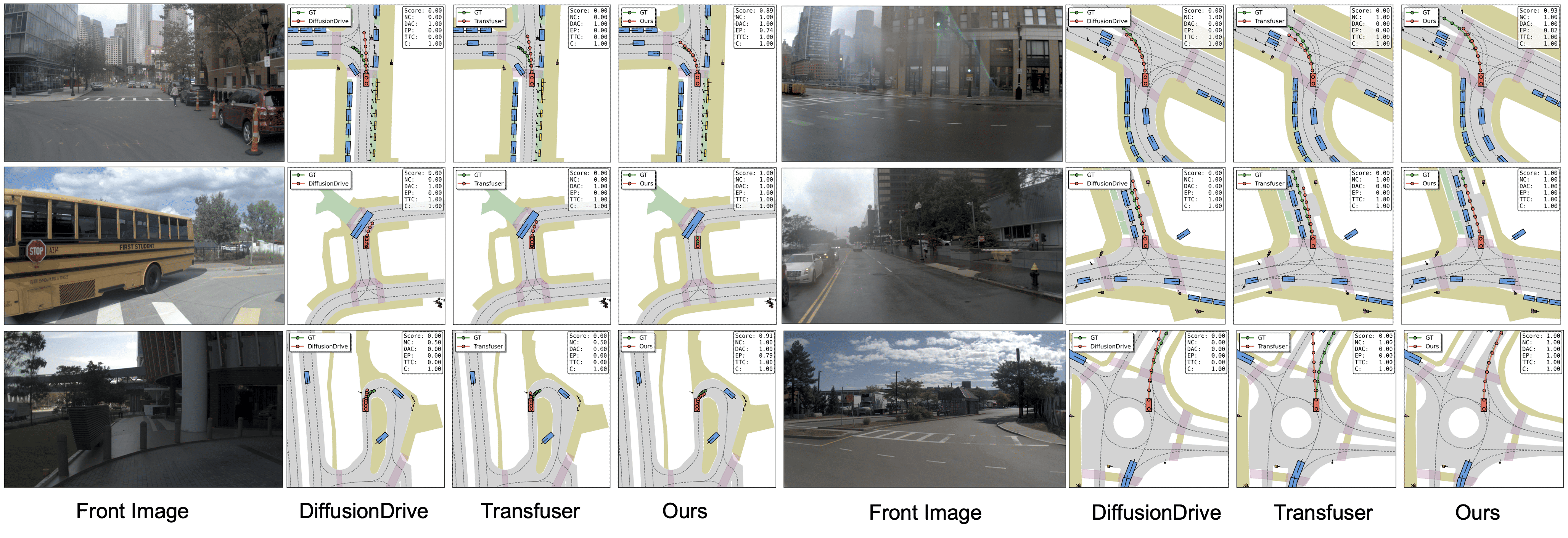} 
  \vspace{-7mm}
  \caption{Qualitative comparison on NAVSIM.}
  \label{fig:navsim_comp}
\end{figure*}

\begin{table}[t!]
\vspace{1mm}
\centering
\small
\renewcommand{\arraystretch}{1.2}
\begin{adjustbox}{max width=\linewidth}
\begin{tabular}{l|ccccccccc|c}
\toprule
\textbf{Method} & \textbf{NC}$\uparrow$ & \textbf{DAC}$\uparrow$ & \textbf{DDC}$\uparrow$ & \textbf{TLC}$\uparrow$ & 
\textbf{EP}$\uparrow$ & \textbf{TTC}$\uparrow$ & \textbf{LK}$\uparrow$ & \textbf{HC}$\uparrow$ & \textbf{EC}$\uparrow$ & \textbf{EPDMS}$\uparrow$ \\
\midrule
Ego Status & 93.1 & 77.9 & 92.7 & 99.6 & 86.0 & 91.5 & 89.4 & \textbf{98.3} & 85.4 & 64.0 \\
VADv2~\cite{VADv2-2024} & 97.3 & 91.7 & 98.2 & 99.7 & 77.6 & 92.7 & 66.0 & \textbf{98.3} & 83.3 & 76.6 \\
TransFuser~\cite{Transfuser-2022} & 97.7 & 92.8 & 98.3 & 99.7 & 79.2 & 92.8 & 67.6 & \textbf{98.3} & \textbf{87.2} & 77.8 \\
HydraMDP++~\cite{Hydra-MLP-pp-2025} & 97.2 & \textbf{97.5} & 99.4 & 99.6 & 83.1 & 96.5 & 94.4 & 98.2 & 70.9 & 81.4 \\
Artemis~\cite{Artemis-2025} & 98.3 & 95.1 & 98.6 & \textbf{99.8} & 81.5 & \textbf{97.4} & 96.5 & \textbf{98.3} & - & 83.1 \\
RecogDrive~\cite{RecogDrive-2025} & 98.3 & 94.2 & 98.8 & \textbf{99.8} & 86.5 & 97.3 & 96.8 & \textbf{98.3} & 87.7 & 83.6 \\
\midrule
\textbf{Ours} & \textbf{98.5} & 94.5 & \textbf{99.5} & \textbf{99.8} & \textbf{86.9} & 96.8 & \textbf{97.4} & 97.6 & 73.9 & \textbf{84.7}\\
\bottomrule
\end{tabular}
\label{tab:navsim_results}
\end{adjustbox}
\vspace{-1mm}
\caption{Comparison on NAVSIM-v2 with extended metrics.
}
\label{tab:navsim_v2}
\end{table}

\begin{table}[t!]
\centering
\renewcommand{\arraystretch}{1.1}
\begin{adjustbox}{max width=\linewidth}
\begin{tabular}{cccccccccc}
\toprule
\begin{tabular}[c]{@{}c@{}}\textbf{Numerical} \\ \textbf{Tokenizer}\end{tabular}
& \begin{tabular}[c]{@{}c@{}}\textbf{Metric} \\ \textbf{-aligned}\end{tabular} 
& \begin{tabular}[c]{@{}c@{}}\textbf{Pre-} \\ \textbf{training}\end{tabular}
& \begin{tabular}[c]{@{}c@{}}\textbf{SG} \\ \textbf{GRPO}\end{tabular}
& NC$\uparrow$ & DAC$\uparrow$ & TTC$\uparrow$ & Comf.$\uparrow$ & EP$\uparrow$ & \textbf{PDMS}$\uparrow$ \\
\midrule
\ding{55} & \ding{55} & \ding{55} & \ding{55} & 95.8 & 87.5 & 88.6  & 99.5 & 71.7 & 76.2 \\
\ding{51} & \ding{55} & \ding{55} & \ding{55} & 97.0  &91.3  &91.0   &98.9 & 76.4  & 81.1 \\
\ding{51} & \ding{51} & \ding{55} & \ding{55} & 97.4  & 92.6  & 95.3   & 99.3  & 77.5 & 83.4 \\
\ding{51} & \ding{51} & \ding{51} & \ding{55} & 98.5  & 95.1  & 94.4  & 99.5 & 81.8 & 86.7 \\
\ding{51} & \ding{51} & \ding{55} & \ding{51} & 98.4 & 96.1 & 95.3 & 99.5 & 79.3 & 86.9   \\
\ding{51} & \ding{51} & \ding{51} & \ding{51} & \textbf{99.2} & \textbf{98.3} & \textbf{97.0} & \textbf{99.7} & \textbf{82.3} & \textbf{90.3}   \\
\bottomrule
\end{tabular}
\end{adjustbox}
\caption{
Ablation study for the proposed components. We evaluate the effect of metric-aligned numerical tokenizer, VQA pretraining and simulator-guided GRPO on NAVSIM-v1. Row~1 uses the text tokenizer from Janus-1.5B to tokenize the number. ``SG GRPO'' refers to ``Simulator-Guided GRPO''.
}
\label{tab:ablation_main}
\end{table}

\begin{figure*}[t!]
  \centering
  \includegraphics[width=\linewidth]{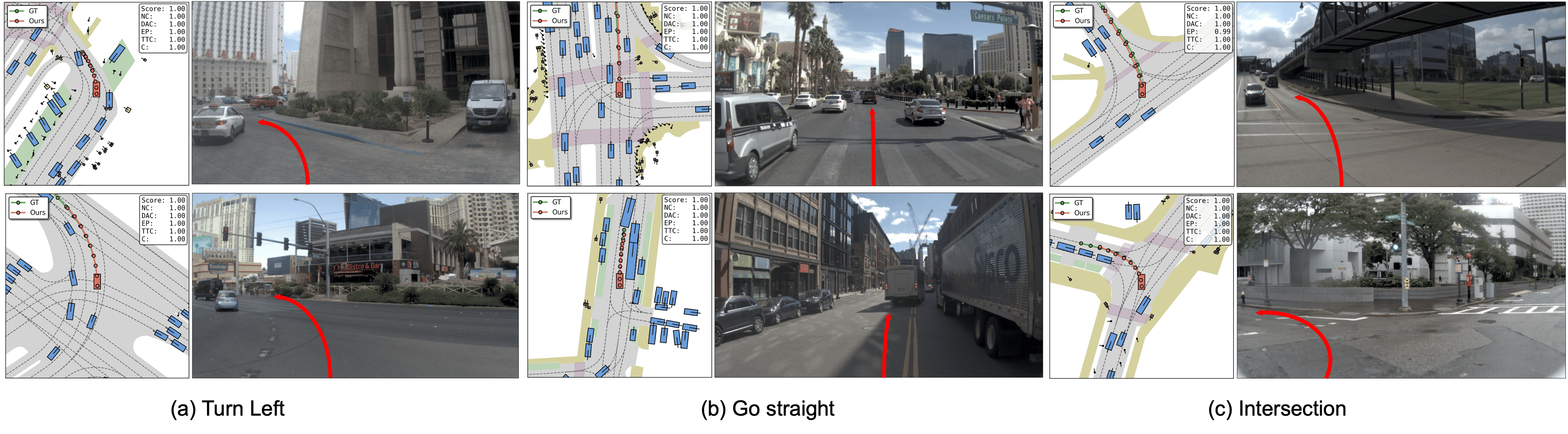}
  \vspace{-6mm}
  \caption{Qualitative results of \model\space on NAVSIM with different scenes.}
  \label{fig:navsim_vis}
\end{figure*}

\section{Experiments}
\subsection{Experimental Setup} 
\noindent\textbf{Implementation.}
All experiments were conducted on 4$\times$8 Ascend 910B NPUs across four sequential training phases. We use AdamW optimizer for all training stages with weight decay of 0.01. In the metric-aligned numerical embeddings training stage, we set $\alpha$ to 0.05, constant learning rate to $1\times10^{-5}$ and batch size to 80. In the pre-training stage, we set constant learning rate to $1\times10^{-5}$ and batch size to 256. In the SFT stage, we utilize learning rate of $5\times10^{-6}$ with cosine annealing strategy and batch size of 64. In the reinforcement learning stage, we use a learning rate of $1\times10^{-6}$, batch size of 32 and 500 warm-up steps. During inference, we use a timestep schedule defined as $\beta_{t} = 3 \times \left(\frac{t}{1 - t}\right)^{0.9}$, and perform inference with 1, 2, 3, 5, and 10 sampling steps. On NAVSIM-v1 benchmark, we conduct evaluations using the v1.1 version of the NAVSIM codebase, while on NAVSIM-v2 benchmark, we use the v2.2 version for evaluation.

\noindent\textbf{Metrics.}
We evaluate our method on the closed-loop NAVSIM-v1~\cite{NAVSIM-v1} and v2~\cite{NAVSIM-v2} benchmarks. 
The primary metric for NAVSIM-v1 is the Predictive Driver Model Score (PDMS), 
a composite measure integrating five key components: 
No-Collision rate (NC), 
Drivable Area Compliance (DAC), 
Time-to-Collision within bound (TTC), 
Comfort, 
and Ego Progress (EP). 
The more comprehensive NAVSIM-v2 benchmark employs the Extended Predictive Driver Model Score (EPDMS), 
which incorporates nine sub-metrics—NC, DAC, Driving Direction Compliance (DDC), 
Traffic Light Compliance (TLC), EP, TTC, Lane Keeping (LK), History Comfort (HC), and Extended Comfort (EC)--
to provide a holistic assessment of driving performance, safety, and rule adherence. 
All results are obtained from closed-loop simulations on the official public test splits.

\subsection{Comparison with State-of-the-Art}
\noindent\textbf{NAVSIM-v1.}
As shown in Table~\ref{tab:navsim_v1}, 
our method achieves the highest PDMS (90.3) on the NAVSIM v1 benchmark. 
It attains the superior performance in both safety-critical metrics--No-Collision (NC: 99.2) and Drivable Area Compliance (DAC: 98.3)--demonstrating superior safety and rule adherence. 
Notably, despite utilizing only a single front-view camera, 
our model outperforms methods that rely on multi-view camera setups or LiDAR inputs, 
underscoring the efficacy of the discrete flow matching paradigm in achieving robust and efficient planning. 
Qualitative analyses (Figure~\ref{fig:navsim_comp} and Figure~\ref{fig:navsim_vis}) further illustrate that our planner produces stable, 
human-like trajectories in closed-loop simulation.

\noindent\textbf{NAVSIM-v2.}
Table~\ref{tab:navsim_v2} presents the evaluation results on the more comprehensive NAVSIM-v2 benchmark. 
Our method achieves the superior overall EPDMS of 84.7. 
It also leads in several critical sub-metrics: 
No-Collision (NC: 98.5), 
Driving Direction Compliance (DDC: 99.5), 
and Lane Keeping (LK: 97.4). 
The superior performance across diverse and dynamic scenarios underscores the robustness of our approach for reliable closed-loop driving.

\begin{figure}[t!]
  \vspace{0.2cm}
  \centering
  \includegraphics[width=0.8\linewidth]{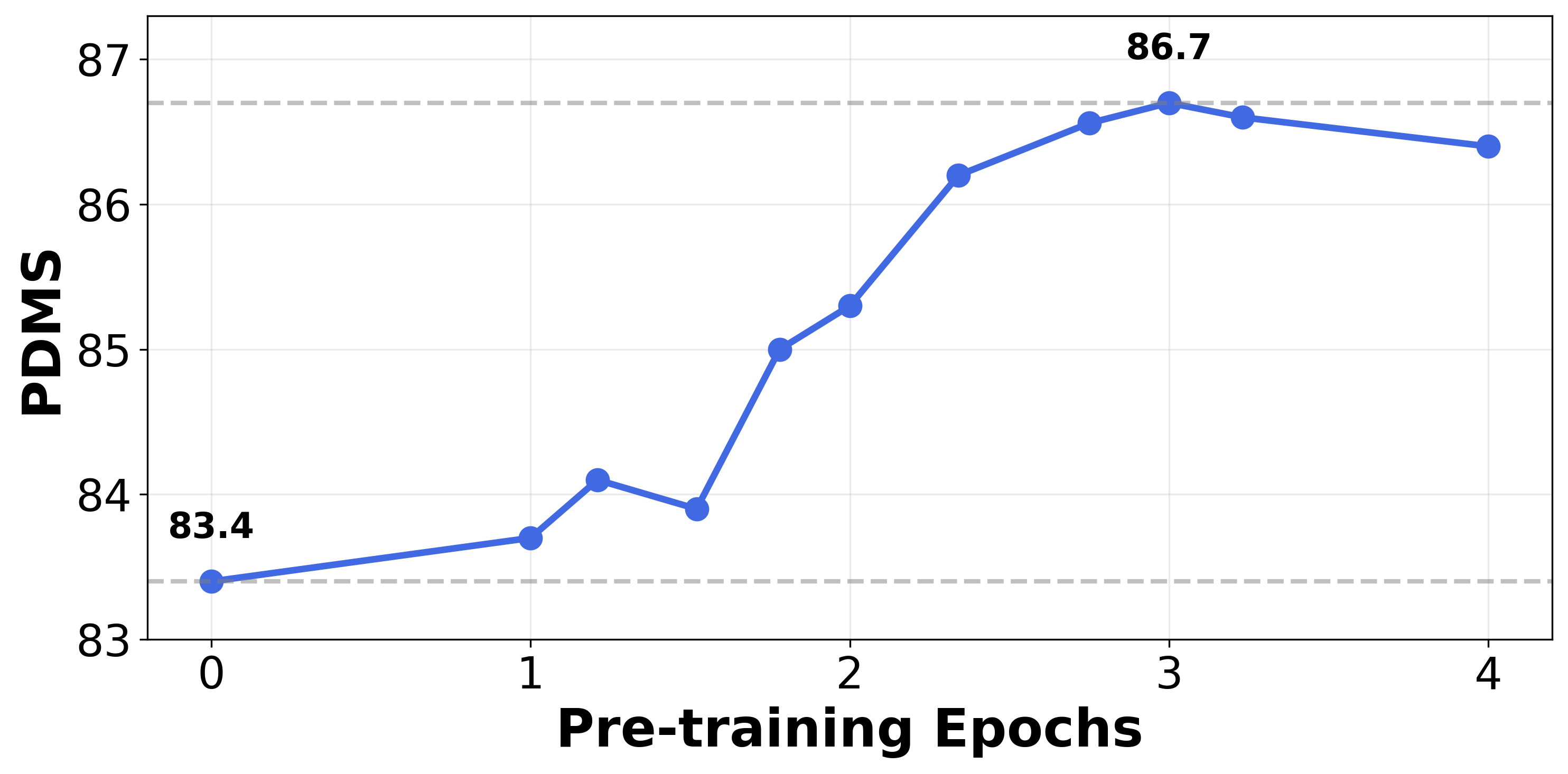}
  \vspace{-0.2cm}
  \caption{Impact of pre-training epochs. We perform SFT after pre-training on 6.5M data, and then calculate PDMS.}
  \label{fig:ablation_pretrain_epoch}
  \vspace{0.2cm}
\end{figure}

\subsection{Ablation and Discussion}
\begin{figure}[t!]
  \centering
  \includegraphics[width=0.8\linewidth]{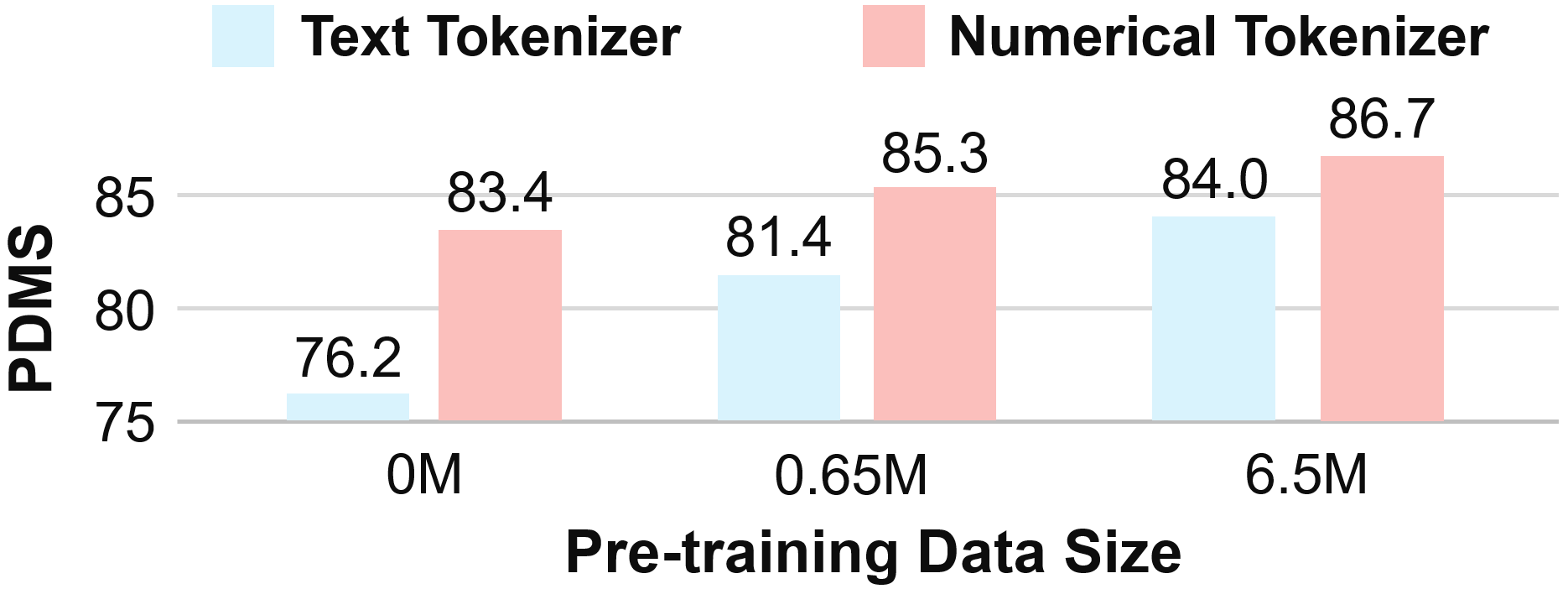}
  \vspace{-2mm}
  \caption{Effect of pretraining dataset scale.}
  \vspace{-3mm}
  \label{fig:ablation_pretrain_data}
\end{figure}



\begin{figure}[t!]
  \centering
  \includegraphics[width=\linewidth]{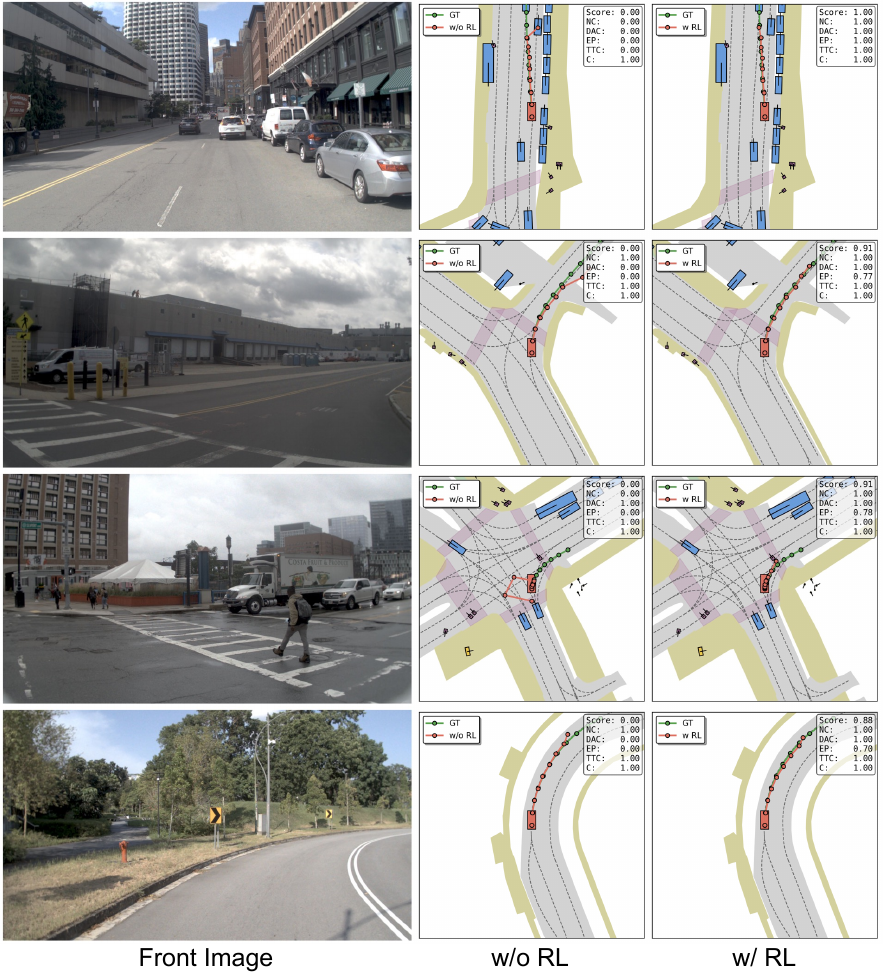} 
  \vspace{-5mm}
  \caption{Ablation about Simulator-guided GRPO.}
  \label{fig:ablation_grpo}
\end{figure}

\noindent\textbf{Ablation Study for Proposed Components.}
As shown in Table~\ref{tab:ablation_main}, 
we systematically evaluate the contribution of each component in our framework. 
Using the Janus-1.5B text tokenizer for numerical values results in a PDMS of 76.2, 
indicating its inadequacy for representing fine-grained trajectory data. 
Replacing it with a dedicated numerical tokenizer improves PDMS by 4.9 points (to 81.1), 
confirming the necessity of a specialized numerical representation. 
Further incorporating metric-aligned embeddings yields an additional gain of 2.3 points (to 83.4), 
demonstrating that geometric consistency in the token space enhances planning quality. 
Subsequent large-scale VQA pretraining adds 3.3 points (to 86.7), 
underscoring the benefit of cross-modal domain adaptation. 
Finally, integrating simulator-guided GRPO achieves the highest PDMS of 90.3, highlighting the critical role of safety and performance alignment through online reinforcement learning. 

In addition, comparing Rows~5 and~6 in Table~\ref{tab:ablation_main} reveals that incorporating VQA pre-training yields a +3.4 improvement in PDMS, 
underscoring the efficacy of pre-training in enhancing driving performance. 
This gain demonstrates that domain-specific pre-training on large-scale visual question answering data provides valuable foundational knowledge for complex driving scenarios, 
complementing the benefits of reinforcement learning-based fine-tuning.


\noindent\textbf{Further Pretrain on More Driving Data.}
Figure~\ref{fig:ablation_pretrain_epoch} illustrates the impact of pre-training epochs on model performance. For a fair comparison, we conduct SFT following pre-training on 6.5M VQA dataset. The results show that the PDMS score increases with the number of pre-training epochs, reaching a peak at 3 epochs, with an improvement of +3.3 compared to 0 epochs. This indicates that pre-training on driving-related VQA tasks significantly enhances the model's driving capabilities in road scenarios.

Figure~\ref{fig:ablation_pretrain_data} investigates the scaling laws of pre-training data. We found that pre-training with 0.65M data yielded a PDMS improvement of +1.9 (+5.2) for numerical (text) tokenizer. Further pre-training with 6.5M data resulted in an additional PDMS increase of +1.4 (+2.6) for numerical (text) tokenizer compared to the 0.65M data. These results highlight the necessity of further pre-training using driving VQA data and confirm the validity of the data scaling law in \model.

\noindent\textbf{Different Simulator-Guided GRPO Settings.}
We analyze the impact of two key design choices in our simulator-guided GRPO framework: group size and reward weighting. 
As shown in Table~\ref{tab:ablation_group_size}, 
varying the group size (number of candidate trajectories sampled per context) reveals a clear trade-off. 
While smaller groups (size=2) yield marginal gains, 
a group size of 3 achieves the optimal balance between exploration diversity and training stability, 
producing the highest PDMS (90.3). 
Larger groups (size=4) introduce excessive variance, slightly degrading performance. 

We further examine the reward function's component weights (EP:TTC:Comfort) in Table~\ref{tab:ablation_reward}. 
Extreme weightings—over-prioritizing either safety (5:20:2) or progress (20:5:2)--
suboptimally skew the policy, 
whereas a balanced ratio (5:5:2) best harmonizes these competing objectives, 
achieving the superior PDMS of 90.3. 
This indicates that equitable consideration of ego progress, safety, and comfort is crucial for well-rounded driving performance.



\begin{table}[t!]
\vspace{2mm}
\centering
\small
\renewcommand{\arraystretch}{1.05}
\begin{adjustbox}{max width=\linewidth}
\begin{tabular}{lccccc}
\toprule
\textbf{Method} & \textbf{Paradigm}  & \textbf{Step} & \textbf{PDMS}$\uparrow$ & \textbf{Infer Time}$\downarrow$ & \textbf{Backbone} \\
\midrule
FSDrive~\cite{FutureSightDrive-2025} & AR & - & 85.1 &   10.58s  & Qwen2-VL-2B~\cite{wang2024qwen2}  \\
Epona~\cite{Epona-2025} & AR + Diff.  & - & 86.2 &  1.24s & DiT-2.5B~\cite{DiT-2023} \\
ReCogDrive~\cite{RecogDrive-2025} & AR + Diff.  & - & 89.6 & 0.42s & InternVL3-8B~\cite{InternVL3-2025} \\
Janus-1.5B~\cite{Janus-2025} & AR  & - & - & 0.27s  & - \\  
\midrule
\multirow{5}{*}{\textbf{Ours}}
& \multirow{5}{*}{DFM}
 & 1 & 89.1 & \textbf{0.09s} & \multirow{5}{*}{Janus-1.5B~\cite{Janus-2025}} \\ 
& & 2 & 89.7 & 0.19s &  \\ 
&  & 3 & 90.0 & 0.29s &  \\ 
&  & 5 & \textbf{90.3} & 0.48s & \\ 
& & 10 & 90.2 & 0.94s & \\
\bottomrule
\end{tabular}
\end{adjustbox}
\vspace{-2mm}
\caption{Intuitive efficiency analysis on NAVSIM.}
\vspace{-2mm}
\label{tab:coarse-to-fine}
\end{table}

\noindent\textbf{Coarse-to-fine Sampling Analysis.}
Table~\ref{tab:coarse-to-fine} analyzes the coarse-to-fine property of \model\space by varying the number of parallel denoising steps during inference. 
Increasing the sampling steps from 1 to 5 yields a monotonic improvement in PDMS (89.1 to 90.3), 
demonstrating that iterative refinement enhances planning quality. 
Inference time scales approximately linearly with the number of steps, 
reflecting the parallel nature of the discrete flow matching process. 
This establishes a flexible trade-off: 
fewer steps enable faster, 
coarser plans suitable for real-time constraints, 
while more steps produce higher-fidelity trajectories.

\section{Conclusion}
We present \model, 
a vision–language–action model that formulates motion planning as discrete flow matching over a structured token space. 
The framework incorporates a metric-aligned numerical tokenizer to preserve geometric coherence and employs simulator-guided GRPO to enforce safety and performance in closed-loop control. 
Evaluated on NAVSIM-v1 and v2 benchmarks, 
\model\space achieves competitive results,
demonstrating its ability to generate high-quality trajectories with a flexible trade-off between inference speed and planning fidelity. 
This work underscores the potential of discrete flow matching for building reliable and scalable autonomous driving systems.

\clearpage
{
    \small
    \bibliographystyle{ieeenat_fullname}
    \bibliography{main}
}

\clearpage
\appendix
\maketitlesupplementary

This appendix provides additional experimental results and implementation details to complement the main paper. 
Specifically, Section~\ref{sec:app_Additional_Experiments} presents extended evaluations on the nuScenes~\cite{nuScenes-dataset-2020} datasets, 
along with additional qualitative experiments on NAVSIM. 
Section~\ref{sec:app_Pseudocode} provides pseudocode for the training and inference stages, respectively.
Section~\ref{sec:app_metrics} elaborates on the evaluation metrics, and Section~\ref{sec:app_implementation_details} discusses implementation specifics.
Finally, Section~\ref{sec:limitation} discuss the limitation and future work.

\section{Additional Experiments}
\label{sec:app_Additional_Experiments}

\subsection{nuScenes Results}
We evaluate our method on the nuScenes dataset~\cite{nuScenes-dataset-2020} following the NAVSIM benchmark perspective~\cite{NAVSIM-v1, NAVSIM-v2}, 
which focuses on collision rate as the primary metric.
This emphasis stems from the established finding in NAVSIM that open-loop L2 distance exhibits negligible correlation with closed-loop performance.
As shown in Table~\ref{tab:Nuscenes}, 
our method achieves an average collision rate of \textbf{0.12\%} under ST-P3 metrics, 
matching the performance of the best non-VLA model (UniAD).
More notably, under the more comprehensive UniAD metrics, 
\model\space sets a new state-of-the-art with the lowest average collision rate (\textbf{0.23\%}) among all evaluated VLA methods. 
The model also demonstrates superior short-term safety, 
achieving a perfect \textbf{0.00\%} collision rate at the 1-second horizon. 

\begin{table*}[h!]
    \centering
    \setlength{\tabcolsep}{10pt}
    \resizebox{\linewidth}{!}{ 
    \renewcommand{\arraystretch}{1.1}
    \begin{tabular}{lcccccccccc}
    \toprule
     \multirow{4}{*}{\textbf{Method}} & \multirow{4}{*}{{\textbf{Paradigm}}} & \multirow{4}{*}{{\textbf{Backbone}}} & \multicolumn{8}{c}{\textbf{Collision (\%) ↓}} \\
    \cmidrule(lr){4-11}
    &  &  & \multicolumn{4}{c}{\textbf{ST-P3 metrics}} & \multicolumn{4}{c}{\textbf{UniAD metrics}} \\
    \cmidrule(lr){4-7} \cmidrule(lr){8-11}
    &  &  & 1s & 2s & 3s & \cellcolor{gray!20}Avg. & 1s & 2s & 3s & \cellcolor{gray!20}Avg. \\
    \midrule
    \multicolumn{11}{l}{\textit{End-to-End}} \\
    PreWorld~\cite{Preworld-2025} & - & - & - & - & - & - & 0.19 & 0.57 & 2.65 & 1.14 \\ 
    ST-P3~\cite{ST-P3-2022} & - & - & 0.23 & 0.62 & 1.27 & 0.71 & - & - & - & - \\ 
    Ego-MLP~\cite{Ego-MLP-2024} & - & - & 0.21 & 0.35 & 0.58 & 0.38 & - & - & - & - \\
    InsightDrive~\cite{InsightDrive-2025} & - & - & 0.09 & 0.10 & 0.27 & 0.15 & 0.08 & 0.15 & 0.84 & 0.36 \\
    VAD-v2~\cite{VADv2-2024} & - & - & 0.07 & 0.10 & 0.24 & 0.14 & - & - & - & - \\
    UniAD~\cite{UniAD-2023} & - & - & \textbf{0.04} & \textbf{0.08} & \textbf{0.23} & \textbf{0.12} & 0.05 & 0.17 & 0.71 & 0.31 \\
    \midrule
    \multicolumn{11}{l}{\textit{End-to-End VLA}} \\
    Epona~\cite{Epona-2025} & AR + Diff. & DiT-2.5B~\cite{DiT-2023} & 0.05 & 0.22 & 0.85 & 0.96 & - & - & - & - \\
    OmniDrive~\cite{OminiDrive-2025} & AR & LLaVA-7B~\cite{LLaVA} & \textbf{0.04} & 0.46 & 2.32 & 0.94 & - & - & - & - \\
    DriveVLM~\cite{tian2024drivevlm} & AR & Qwen2-VL-7B~\cite{wang2024qwen2} & 0.10 & 0.22 & 0.45 & 0.27 & - & - & - & - \\
    GPT-Driver \cite{GPT-Driver-2023} & AR & GPT-4~\cite{GPT-4-2023} & 0.04 & 0.12 & 0.36 & 0.17 & 0.07 & 0.15 & 1.10 & 0.44 \\
    AutoVLA~\cite{zhou2025autovla} & AR & Qwen2.5-3B~\cite{Qwen-25} & 0.13 & 0.18 & 0.28 & 0.20 & 0.14 & 0.25 & \textbf{0.53} & 0.31 \\
    DME-Driver~\cite{DME-Driver-2025} & AR & LLaVA-7B~\cite{LLaVA} & - & - & - & - & 0.05 & 0.28 & 0.55 & 0.29 \\
    \textbf{Ours} & DFM & Janus-1.5B~\cite{Janus-2025} & \textbf{0.04} & 0.10 & \textbf{0.23} & \textbf{0.12} & \textbf{0.00} & \textbf{0.10} & 0.60 & \textbf{0.23} \\
    \bottomrule
    \end{tabular}}
    \vspace{-2mm}
    \caption{
        End-to-end motion planning performance on the nuScenes~\cite{nuScenes-dataset-2020} dataset. We sort previous methods according to the average collision rate. Abbreviation: Diff.(Diffusion), AR (autoregressive), DFM (discrete flow matching).
    }
    \label{tab:Nuscenes}
    \vspace{3mm}
\end{table*}

\begin{table*}[ht!]
\centering
\resizebox{\linewidth}{!}{
\begin{tabular}{@{}ccccc@{}}
\toprule
 \multirow{2}{*}{\textbf{Hyperparameter}} & \textit{\textbf{Stage 1}} & \textit{\textbf{Stage 2}} & \textit{\textbf{Stage 3}} & \textit{\textbf{Stage 4}} \\ \cmidrule(l){2-5} 
 & \textbf{Embedding Training} & \textbf{Pre-training} & \textbf{Supervised Fine-tuning} & \textbf{Reinforcement Learning} \\ \midrule

Training Modules & Numerical Tokenizer & VLA & VLA & VLA \\
Training Parameters & 0.4B & 1.5B & 1.5B & 1.5B \\

Training Data & nuPlan (668K) & VQA (6.5M) & nuPlan (668K) & NAVSIM (103K) \\
Loss & $\mathcal{L}_{\mathrm{CE}} + \mathcal{L}_{\mathrm{num}}$ & $\mathcal{L}_{\mathrm{CE}}$ & $\mathcal{L}_{\mathrm{CE}}$ & $\mathcal{L}_{\mathrm{GRPO}}$ \\

Training Epochs & 4 & 3 & 2 & 0.5 \\
Batch Size & 80 & 256 & 64 & 32 \\
Optimizer & Adam & Adam & Adam & Adam \\
Learning Rate & $1\times10^{-5}$ & $1\times10^{-5}$ & $5\times10^{-6}$ & $1\times10^{-6}$ \\
Learning Rate Scheduler & constant & constant & cosine annealing & cosine annealing \\
Warm-up Steps & 0 & 0 & 500 & 500 \\ 
Gradient Accumulation Steps & 1 & 1 & 1 & 1 \\
\bottomrule
\end{tabular}
}
\vspace{-2mm}
\caption{
Key hyperparameters for different training stages.
}
\label{tab:app_training_hyper_parameter}
\end{table*}

\subsection{NAVSIM Qualitative Results}
\begin{figure*}[t!]
  \centering
  \includegraphics[width=\textwidth]{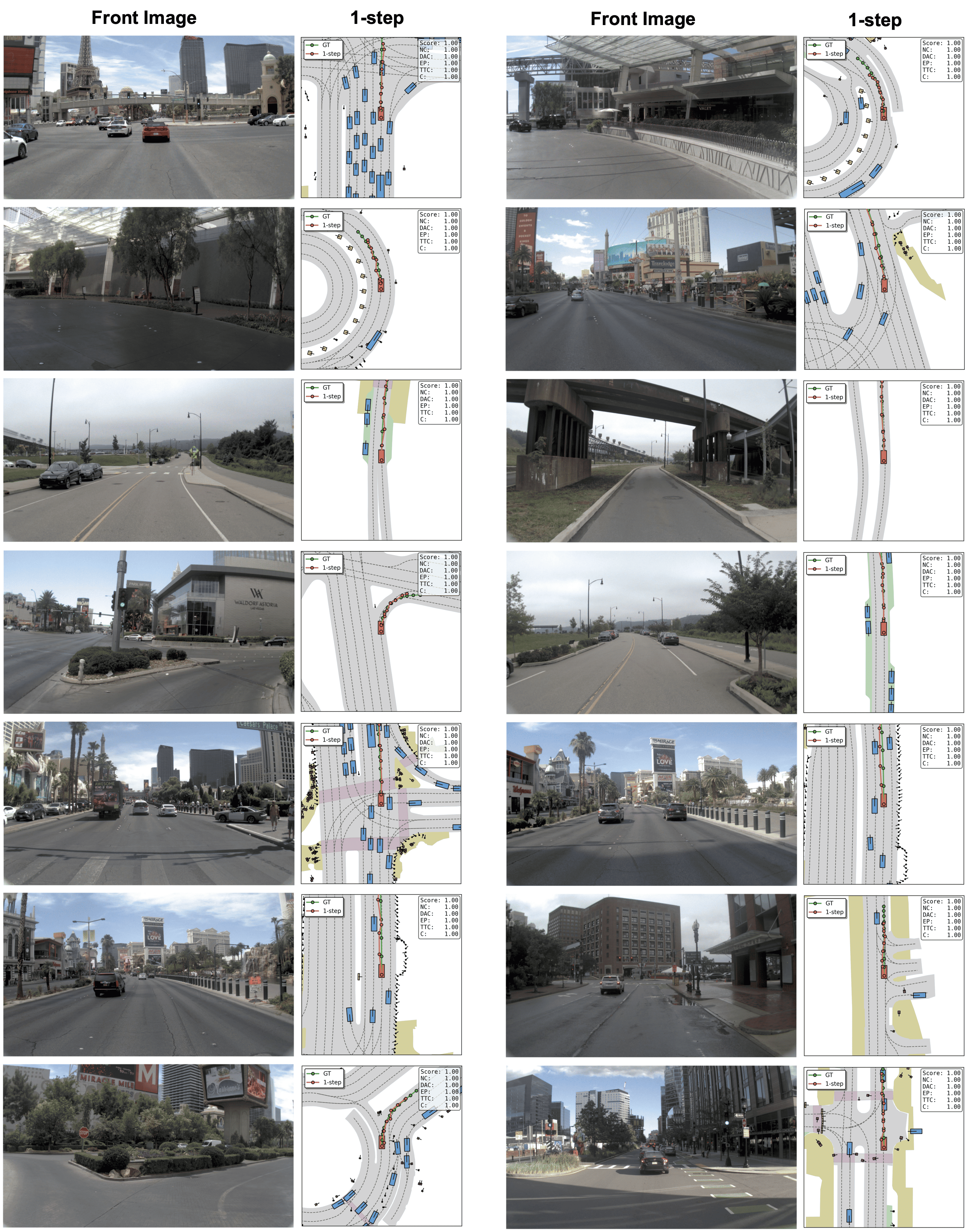}
  \caption{
     For straightforward driving scenarios on NAVSIM, our method achieves acceptable outcomes with just 1-step denoising.
  }
  \label{fig:app_vis_navsim_easy}
\end{figure*}

\begin{figure*}[t!]
  \centering
  \includegraphics[width=0.95\textwidth]{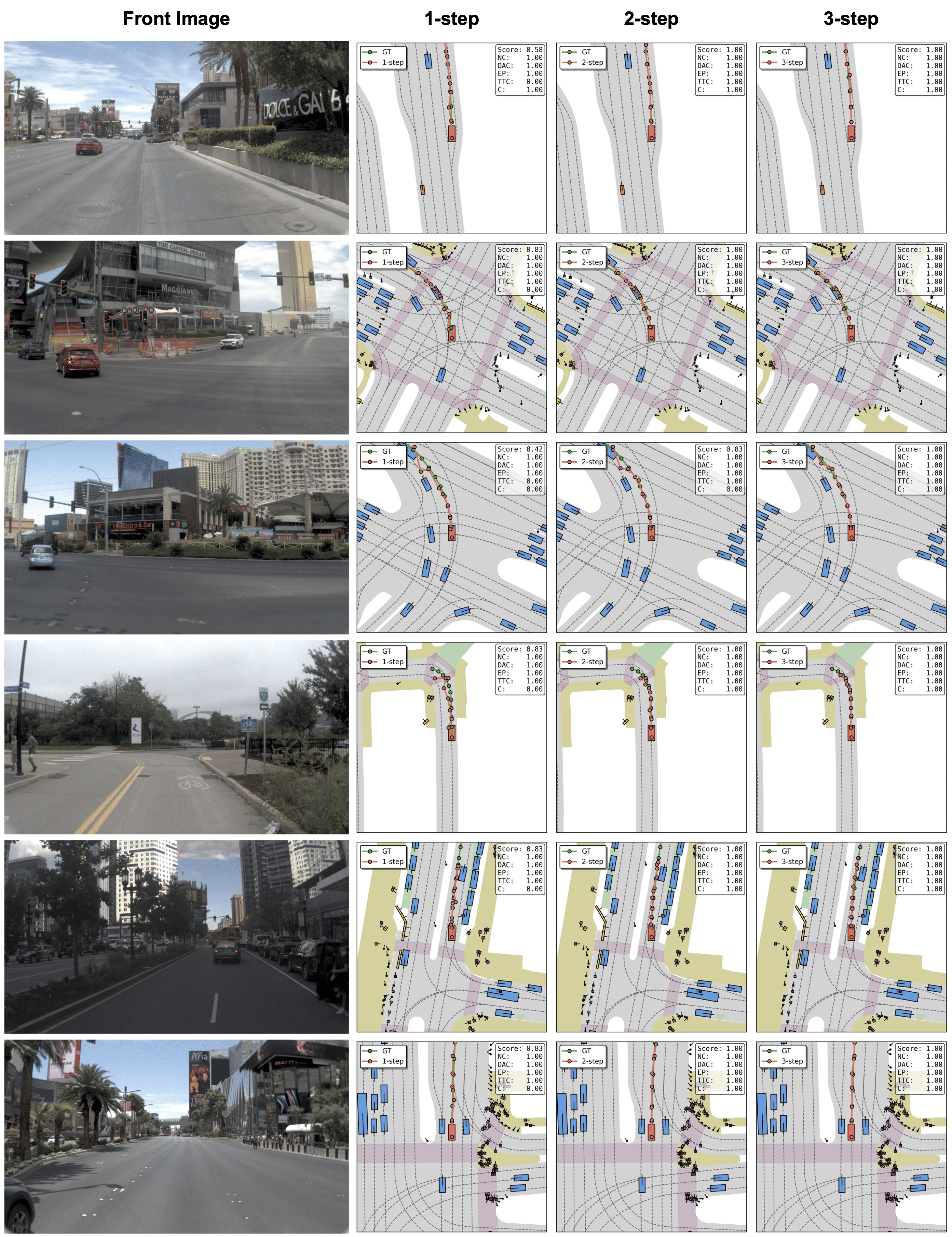}
  \caption{
    Visualization of the 3-step refinement results on NAVSIM.
  }
  \label{fig:app_vis_navsim_medium}
\end{figure*}

\begin{figure*}[t!]
  \centering
  \includegraphics[width=\textwidth]{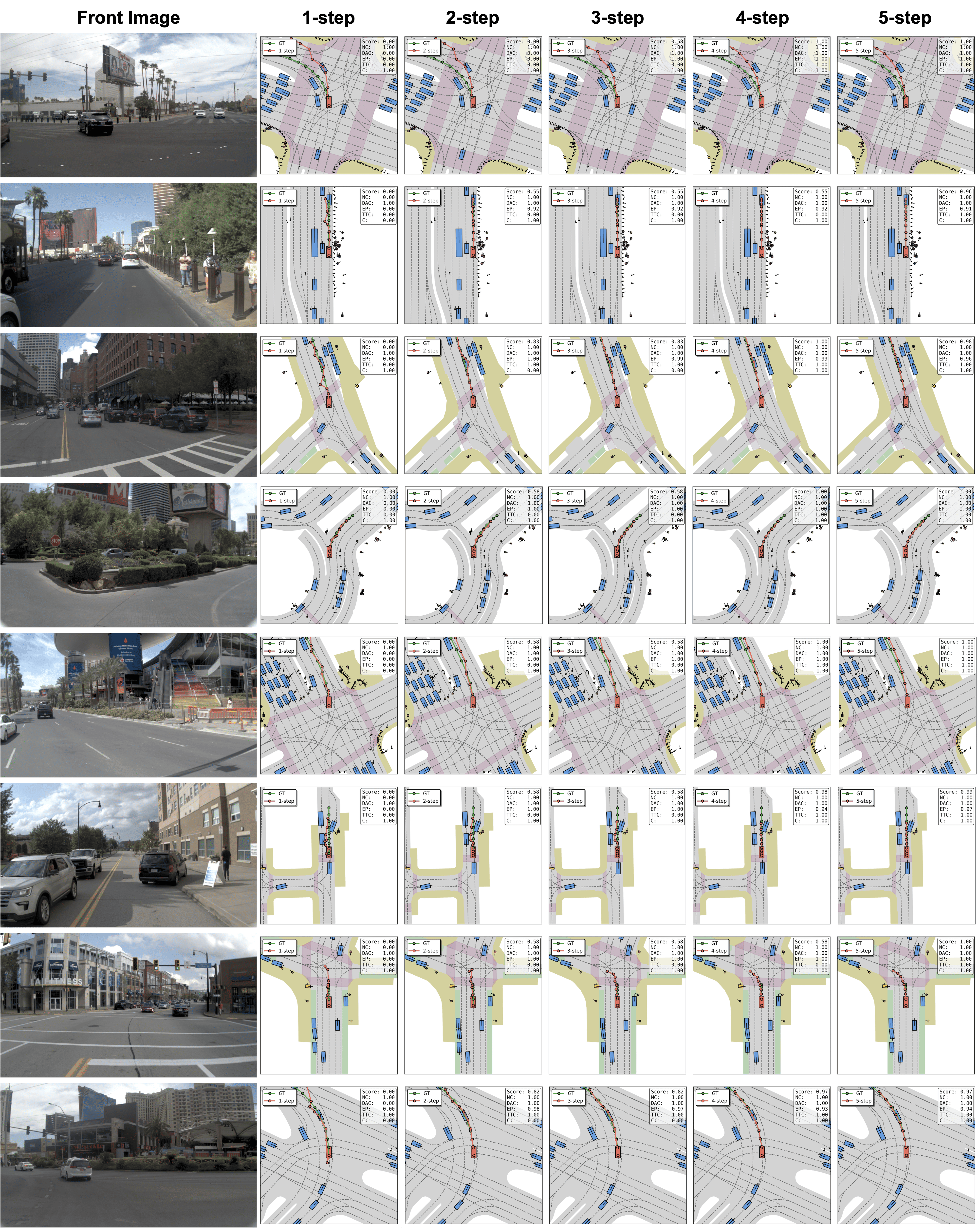}
  \caption{
    For relatively complex scenarios on NAVSIM, our model generates reasonable results through a 5-step coarse-to-fine trajectory prediction process.
  }
  \label{fig:app_vis_navsim_hard}
\end{figure*}

Figure~\ref{fig:app_vis_navsim_easy},~\ref{fig:app_vis_navsim_medium} and ~\ref{fig:app_vis_navsim_hard} visualizes 1-, 3- and 5-step results on NAVSIM, respectively. For straightforward driving scenarios (Figure~\ref{fig:app_vis_navsim_easy}), \model\space generates acceptable trajectories with only 1-step denoising. For relatively complex scenarios (Figure~\ref{fig:app_vis_navsim_hard}), our method predicts reasonable results through a 5-step parallel coarse-to-fine process.

\section{Pseudocode for Training and Inference}
\label{sec:app_Pseudocode}

Algorithm~\ref{alg:cap1} and \ref{alg:cap2} respectively describe the training and inference procedure.

\begin{algorithm}[!t]
\caption{Training}\label{alg:cap1}
{\footnotesize
\begin{algorithmic}[1]
\Require  model parameters $\theta$, time schedule $\beta_t$
\Ensure Optimized parameters $\theta^*$
\State Initialize model parameters $\theta$
\While{not converged}
    \State Sample batch $x_1 \sim q(x)$ \Comment{Trajectory}
    \State Sample $t \sim \mathcal{U}[0,1]$ \Comment{Continuous time sampling}
    \State $p_t(x|x_1) = \mathrm{softmax}(-\beta_t \cdot d(x,x_1))$ \Comment{Compute transition probabilities}
    \State $x_t \sim p_t(x|x_1)$ \Comment{Sample noisy tokens}
    \State $p_{1|t}^{\theta}(\cdot | x_t) = \mathrm{model}_{\theta}(x_t, c)$ \Comment{Compute conditional distribution}
    \State $\mathcal{L}_{\mathrm{CE}} = -\mathbb{E}\left[\sum_{i=1}^{D} \log p_{1|t}^{\theta,i}(x_1^i|x_t)\right]$ \Comment{Compute loss}
    \State Update $\theta$ via gradient descent on $\mathcal{L}_{\mathrm{CE}}$
\EndWhile
\end{algorithmic}}
\end{algorithm}

\begin{algorithm}[!t]
\caption{Inference}\label{alg:cap2}
{\footnotesize
\begin{algorithmic}[1]
\Require Number of inference steps $n$
\Ensure Generated token sequence $x_1$
\State $h \gets 1/n$ \Comment{Step size for Euler discretization}
\State Initialize $x_0$: for each coordinate $i$, sample $x_0^i$ uniformly from vocabulary
\For{$k = 0, 1, \dots, n-1$}
    \State $t \gets k \cdot h$ \Comment{Current time in $[0,1)$}
    \For{$i = 1$ to $D$ \textbf{in parallel}} \Comment{Parallel processing of all coordinates}
        \State Compute posterior: $p_{1|t}^{\theta,i}(\cdot|x_t) \gets \mathrm{model}_{\theta}(x_t, c)$
        \State Sample target: $x_1^i \sim p_{1|t}^{\theta,i}(\cdot|x_t)$
        \State Compute total transition rate: $\lambda_i \gets \sum_{y^i \neq x_t^i} u_t^i(y^i, x_t^i | x_1^i)$
        \State Sample threshold: $Z_i \sim \mathcal{U}[0,1]$
        \If{$Z_i \leq 1 - e^{-h\lambda_i}$} \Comment{Transition occurs with probability $1 - e^{-h\lambda_i}$}
            \State Sample new token: $x_{t+h}^i \sim \frac{u_t^i(\cdot, x_t^i | x_1^i)}{\lambda_i}$ 
        \Else
            \State Retain current token: $x_{t+h}^i \gets x_t^i$
        \EndIf
    \EndFor
    \State Advance time: $x_t \gets x_{t+h}$
\EndFor
\State \Return $x_1$ \Comment{Final denoised token sequence at $t=1$}
\end{algorithmic}}
\end{algorithm}

\section{Detailed Explanation for Metrics}
\label{sec:app_metrics}

This section provides detailed definitions of the evaluation metrics used in our experiments.

\subsection{NAVSIM-v1 Metrics}
For NAVSIM-v1~\cite{NAVSIM-v1}, the primary evaluation metric is the Predictive Driver Model Score (PDMS), 
which integrates five key performance indicators:
\begin{equation}
\mathrm{PDMS} = \mathrm{NC} \times \mathrm{DAC} \times \frac{(5 \times \mathrm{TTC} + 2 \times \mathrm{C} + 5 \times \mathrm{EP})}{12}
\end{equation}
\begin{itemize}
\item \textbf{No at-fault Collision (NC)}: Penalizes collisions based on fault assignment. NC=1 indicates no at-fault collisions, NC=0.5 indicates one fault collision with static objects, and NC=0 indicates multiple fault collisions.
\item \textbf{Drivable Area Compliance (DAC)}: Measures adherence to drivable areas (lanes, parking areas). DAC=1 when the ego bounding box remains entirely within drivable areas, and DAC=0 when any corner exits designated areas.
\item \textbf{Ego Progress (EP)}: Quantifies navigation goal achievement as the ratio of actual progress to a search-based safe upper bound derived from PDM-Closed trajectories. The ratio is clipped to [0,1], with low or negative values discarded.
\item \textbf{Time-to-Collision (TTC)}: Encourages maintenance of safe distances from other vehicles. TTC=1 when the minimum time-to-collision exceeds 0.9 seconds, and 0 otherwise.
\item \textbf{Comfort (C)}: Assesses kinematic constraints including acceleration and jerk. C=1 when all predefined thresholds are satisfied, and 0 upon any violation.
\end{itemize}

\subsection{NAVSIM-v2 Metrics}
For NAVSIM-v2~\cite{NAVSIM-v2}, the Extended Predictive Driver Model Score (EPDMS) incorporates additional safety and compliance measures:
\begin{equation}
\scalebox{0.8}{$
\begin{aligned}
\mathrm{EPDMS} &= \mathrm{NC} \times \mathrm{DAC} \times \mathrm{DDC} \times \mathrm{TL} \times \\
&\quad \frac{(5 \times \mathrm{TTC} + 2 \times \mathrm{C} + 5 \times \mathrm{EP} + 5 \times \mathrm{LK} + 5 \times \mathrm{EC})}{22}
\end{aligned}
$}
\end{equation}
\begin{itemize}
\item \textbf{Driving Direction Compliance (DDC)}: 
Penalizes reverse driving behavior. 
DDC=1 for reverse distance $<2$m, DDC=0.5 for $2-6$m, 
and DDC$=0$ for $>6$m.
\item \textbf{Traffic Light Compliance (TLC)}: 
Measures obedience to traffic signals. 
TLC$=1$ when traffic rules are followed, 
and 0 upon violations.
\item \textbf{Lane Keeping (LK)}: 
Evaluates lateral positioning relative to lane centerlines, 
scored continuously from 0 to 1.
\item \textbf{History Comfort (HC)}: 
Assesses trajectory consistency with historical motion patterns, 
ranging from 0 to 1.
\item \textbf{Extended Comfort (EC)}: 
Compares planned trajectories across consecutive frames for dynamic consistency, 
scored from 0 to 1.
\end{itemize}

\subsection{nuScenes Metrics}
For nuScenes, we follow the NAVSIM~\cite{NAVSIM-v1, NAVSIM-v2} perspective, focusing only on the collision rate. 

\section{Implementation Details}
\label{sec:app_implementation_details}
In Table~\ref{tab:app_training_hyper_parameter}, we show the key hyperparameters for different training steps, including training modules, parameters, data, loss, epochs, batch sizes, optimizer, learning rate, learning rate scheduler, warm-up and gradient accumulation steps.

\section{Limitation and Future Work}
\label{sec:limitation}
While \model\space demonstrates promising results, several limitations warrant attention. 
First, our evaluation is conducted primarily in simulation environments (NAVSIM, nuScenes), 
which may not fully capture the complexities of real-world driving scenarios. 
Second, the GRPO reward is designed for and evaluated in simulation; 
its safety and performance terms require careful redesign to bridge the sim-to-real gap. 
Third, the model is trained and validated on existing benchmarks, 
which may not encompass the full long-tail distribution of real-world driving scenarios.

Future work will explore several directions. 
We plan to extend the framework to support variable-horizon planning and incorporate multi-modal sensor inputs (e.g., LiDAR, radar) for enhanced robustness. 
We also plan to investigate learning a world model as a more generalizable alternative to simulator-based rewards.
Finally, real-world deployment and testing will be essential to validate the model's performance under actual driving conditions.

\end{document}